\documentclass[conference]{IEEEtran}
\usepackage{cite}
\usepackage{amsmath,amssymb,amsfonts}
\usepackage{xcolor}
\usepackage{textcomp}
\usepackage{graphicx}
\usepackage{float}

\renewcommand{\figurename}{Figure}

\def\BibTeX{{\rm B\kern-.05em{\sc i\kern-.025em b}\kern-.08em
    T\kern-.1667em\lower.7ex\hbox{E}\kern-.125emX}}
\begin{document}

\title{Interpretable Credit Default Prediction with Ensemble Learning and SHAP}

\author{\IEEEauthorblockN{1\textsuperscript{st} Shiqi Yang}
\IEEEauthorblockA{\textit{New York University} \\
New York, USA \\
sy3506@nyu.edu}
\and
\IEEEauthorblockN{2\textsuperscript{nd} Ziyi Huang\IEEEauthorrefmark{1}}
\IEEEauthorblockA{\textit{Independent Researcher} \\
Seattle, USA \\
hziyi66@gmail.com\IEEEauthorrefmark{1}}
\and
\IEEEauthorblockN{3\textsuperscript{rd} Wengran Xiao}
\IEEEauthorblockA{\textit{University of Michigan} \\
New York, USA \\
wengranx@umich.edu}
\and
\IEEEauthorblockN{4\textsuperscript{th} Xinyu Shen}
\IEEEauthorblockA{\textit{Georgia Institute of Technology} \\
Atlanta, USA \\
xshen63@gatech.edu}
\thanks{Corresponding Author. Email: alice.johnson@example.com}
}
\maketitle

\begin{abstract}
This study focuses on the problem of credit default prediction, builds a modeling framework based on machine learning, and conducts comparative experiments on a variety of mainstream classification algorithms. Through preprocessing, feature engineering, and model training of the Home Credit dataset, the performance of multiple models including logistic regression, random forest, XGBoost, LightGBM, etc. in terms of accuracy, precision, and recall is evaluated. The results show that the ensemble learning method has obvious advantages in predictive performance, especially in dealing with complex nonlinear relationships between features and data imbalance problems. It shows strong robustness. At the same time, the SHAP method is used to analyze the importance and dependency of features, and it is found that the external credit score variable plays a dominant role in model decision making, which helps to improve the model's interpretability and practical application value. The research results provide effective reference and technical support for the intelligent development of credit risk control systems.
\end{abstract}


\section{Introduction}
In today’s financial landscape, institutions process millions of credit applications every day, making credit-based lending a core function of the modern economic system. Although this activity enables individuals and companies to access funds and drives economic growth, it also faces the significant risk of credit defaults. As the credit business expands, application volumes increase and lending offerings diversify, credit defaults become more frequent, posing serious threats to financial stability and economic development\cite{9240729,8776802}. As a result, accurately and efficiently predicting credit defaults has become a pressing challenge for financial institutions\cite{MOSCATELLI2020113567}.

Traditional credit risk prediction methods primarily rely on statistical analyzes of quantitative variables such as credit utilization and payment history. These approaches generally operate on low-dimensional data and a limited set of predefined factors, which restricts their ability to model complex or nonlinear relationship between diverse attributes and credit default outcomes. In the era of big data, the sources of financial data have become increasingly diverse - including demographic information, transaction records, social behaviors, and geolocation data. While these complex, multi-source, and dynamic data present unprecedented material for enhancing the accuracy of credit default prediction, they also raise significant challenges for traditional statistical models, which lacks the adaptability and scalability to support massive multidimensional data.

Machine learning, a rapidly advancing subfield of artificial intelligence, offers significant potential to overcome the limitations of traditional models. It is capable of processing high-cardinality, nonlinear, and heterogeneous data by extracting features, identifying patterns, and modeling relationships among variables. Moreover, machine learning models can be iteratively refined through techniques such as training on different datasets, applying feature engineering, and optimizing underlying algorithms. These capabilities enable machine learning to uncover hidden relationships behind multidimensional customer data and credit default outcomes, hence improving accuracy and stability of credit default prediction for financial institutions\cite{su12166325}. 

Recent development of emerging technologies such as deep learning, ensemble learning, and transfer learning have further advanced the intelligence and precision of credit default prediction models. For example, Wu et al. demonstrated that deep learning can effectively handle high-dimensional data and generate real-time decisions for robots, showcasing its potential for solving challenging prediction problems in finance as well\cite{wuWarehouse2025}. These advancements improve the scalability, adaptability, and generalization-ability of prediction systems. They have also expanded the technical boundaries of credit risk management and promoted the intelligent transformation and innovative upgrade of financial services\cite{robisco2022,junhui2021}.

Conducting research on credit default prediction algorithms based on machine learning has significant theoretical and practical value. Theoretically, exploring the applicability and performance differences of various machine learning algorithms in credit risk prediction enriches the theoretical system of financial technology as an interdisciplinary field. It also drives algorithm optimization and application innovation. Practically, building efficient, accurate, and interpretable credit default prediction models can significantly enhance the risk control capabilities of financial institutions. It helps reduce non-performing loan rates, optimize resource allocation, and improve market competitiveness. At the same time, reasonable application of prediction models can promote financial inclusion. It can enhance the financial accessibility for small, micro enterprises and marginalized groups. This contributes to economic restructuring and social equity, generating positive social effects\cite{10522232}.

Overall, research on credit default prediction algorithms based on machine learning aligns with the digital transformation of finance and the construction of intelligent risk control systems. It addresses the increasing complexity of financial risk management challenges. With the improvement of computing power and the continuous evolution of algorithmic techniques, credit default prediction will become more accurate, intelligent, and efficient. This study aims to systematically analyze the application effects of different machine learning methods in credit default prediction\cite{DECASTROVIEIRA2019105640}. It also explores strategies for model optimization and performance improvement. The goal is to provide financial institutions with feasible technical solutions and theoretical references, promote the modernization and scientific development of credit risk management systems, and help build a more robust and efficient financial ecosystem.

\section{Method}
In this study, the credit data set is first standardized to improve the weight balance of different features during the training process and avoid model performance fluctuations caused by dimensional differences. The modeling framework is shown in Figure 1.

\begin{figure}[H]
    \centering
    \includegraphics[width=0.5\textwidth]{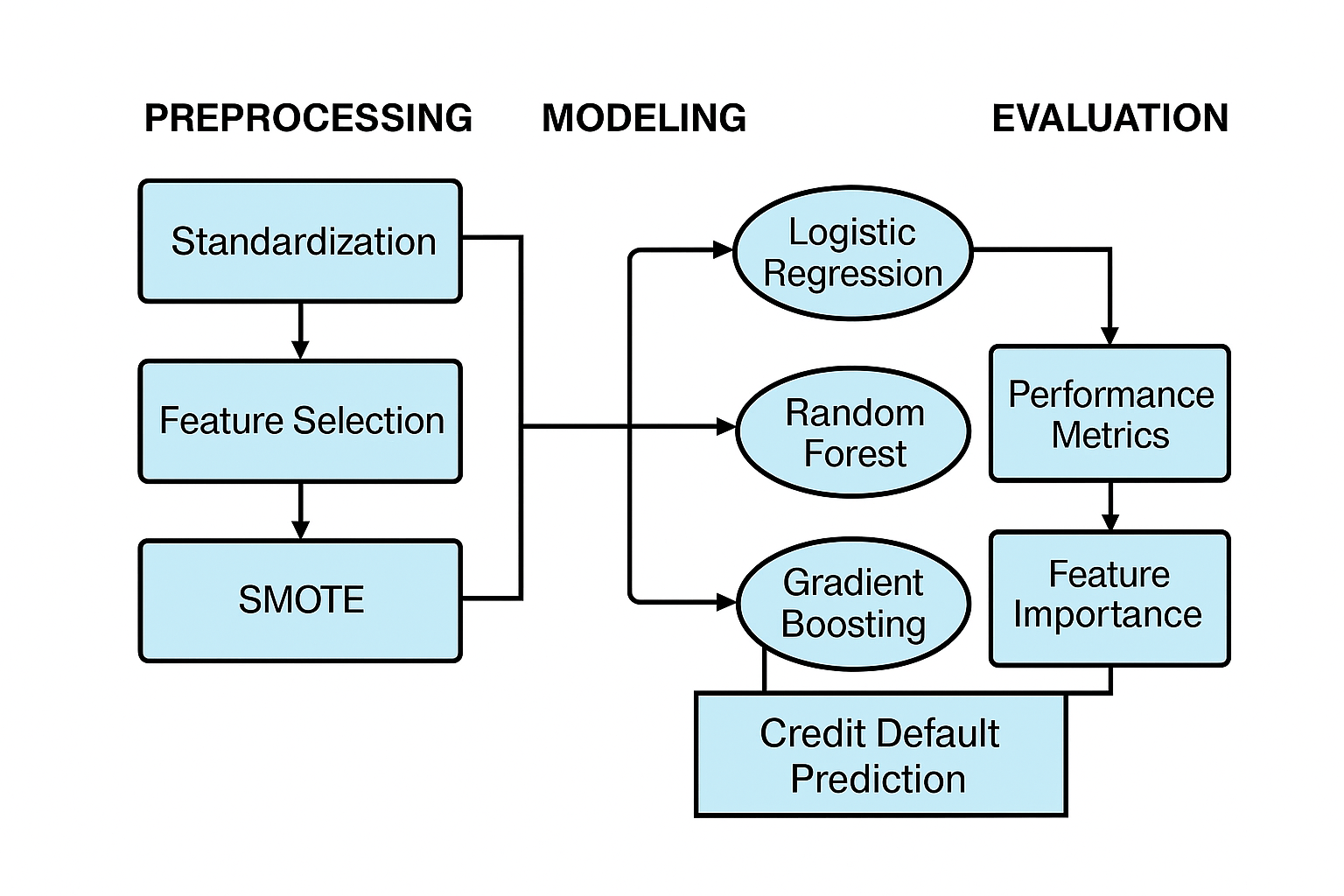}
    \caption{Overall modeling framework diagram.}
\end{figure}

\subsection{Dataset}
The Home Credit dataset is a real-world financial dataset designed for credit risk modeling. It is mainly used to predict whether a loan applicant has a risk of default. The dataset consists of multiple sub-tables, covering the applicant's basic information, credit history, current loan application status, and external behavior data related to the applicant. The main table consists of 307,511 observations, each corresponding to a loan applicant, and 122 features for training. These features provide rich financial and demographic information about the applicant, such as income, occupation, living status, and family structure. Most importantly, the dataset denotes a binary target label, where 1 representing a credit default, and 0 as non-default. This labeling provides a clear target for supervised learning.

In addition, the dataset includes multiple extended tables containing credit card records, previous loan records, POS transaction data, and external scoring information. These data together construct a high-dimensional, multi-time series, and cross-entity information network. For example, the "bureau.csv" table provides an applicant's credit history in other financial institutions, while the "installments\_payments.csv" table records an applicant's historical installment payment behavior. This complex data structure requires the integration of multiple data sources, association matching, and feature construction during the modeling process. As a result, robust feature engineering techniques are essential for effective modeling.

A notable feature of this dataset is the class imbalance, that is, the number of defaulting customers is far less than that of normal customers. This imbalance poses a challenge to model training and requires strategies such as resampling and class weighting to alleviate it. In addition, the data also contains a large number of missing values, outliers, and categorical variables, which requires in-depth cleaning and conversion in the preprocessing stage. Overall, the Home Credit dataset is highly representative of real world customer characteristics.

\subsection{Data Preprocessing}
Assume that the original feature vector is $x$, and the standardized feature is recorded as $z$. The specific conversion formula is as follows:

\begin{equation}
    z={\frac{x-\mu}{\sigma}}
\end{equation}

Among them, $\mu$ represents the feature mean and $\sigma$ represents the feature standard deviation. The standardized data is used as the model input to improve the training convergence speed and classification performance. In the feature engineering stage, effective features are selected through correlation analysis, information gain and principal component analysis to reduce the interference of redundancy and noise on the model. At the same time, in order to solve the problem of class imbalance, the Synthetic Minority Over-sampling Technique (SMOTE) oversampling method is used to enhance the default samples to improve the learning effect of the model on small sample categories.

\subsection{Machine Learning Models}
In the modeling process, this study selected logistic regression, random forest and gradient boosting decision tree as the main machine learning models, and conducted model evaluation and comparison based on the cross-validation method. The logistic regression model aims to maximize the log-likelihood function and construct a binary classification predictor. Its objective function is:

\begin{equation}
    L(\theta)=\sum_{i=1}^{n}\left[y_{i}\log(h_{\theta}(x_{i}))+(1-y_{i})\log(1-h_{\theta}(x_{i}))\right]
\end{equation}

Among them, $h_{\theta}(x_{i})$ represents the predicted probability corresponding to input $x_{i}$, and $y_{i}$ is the actual label. The parameter $\theta$ is iteratively optimized by the gradient descent method to minimize the prediction error. The random forest model improves the stability and anti-overfitting ability of the model by constructing multiple decision trees and performing voting integration. In each round of iteration, the gradient boosting decision tree uses the negative gradient of the current loss function as the fitting target of the new round of weak classifiers, thereby gradually approaching the optimal model.

\subsection{Evaluation}
In the model evaluation stage, the accuracy, recall, F1 score and AUC indicators are used to comprehensively measure the prediction performance. In order to quantify the impact of each feature on the prediction result, a feature importance scoring mechanism based on the change in model output is introduced. The specific calculation formula is as follows:

\begin{equation}
    \mathrm{Feature\ Importance}=\frac{1}{T}\sum_{t=1}^{T}\Delta Loss_{t}
\end{equation}

Among them, $T$ is the total number of decision trees, and $\Delta Loss_{t}$ represents the loss reduction caused by the splitting of a certain feature in the tth tree. Finally, by comparing the performance of different models on the test set, the best algorithm is selected for credit default risk prediction, and the interpretability analysis of the model results is performed to provide decision-making basis and risk warning support for practical applications.

\section{Experiment Results}
In this section, this paper first gives the comparative experimental results of the proposed algorithm and other algorithms, as shown in Table 1.

\begin{table}[h!]
    \centering
    \caption{Model Performance Across Algorithms.}
    \begin{tabular}{|c|c|c|c|}
        \hline
        \textbf{Method} & \textbf{ACC} & \textbf{Precision} & \textbf{Recall} \\
        \hline
        Logistic Regression & 0.7412 & 0.7096 & 0.7031 \\
        Decision Tree & 0.7350 & 0.6928 & 0.7162 \\
        Random Forest & 0.7693 & 0.7551 & 0.7357 \\
        SVM & 0.7358 & 0.7404 & 0.7210 \\
        KNN & 0.7215 & 0.6842 & 0.6938 \\
        Naive Bayes & 0.7079 & 0.6641 & 0.6750 \\
        MLP & 0.7586 & 0.7482 & 0.7295 \\
        AdaBoost & 0.7652 & 0.7510 & 0.7339 \\
        CatBoost & 0.7789 & 0.7665 & 0.7503 \\
        LightGBM & 0.7797 & 0.7689 & 0.7496 \\
        XGBoost & 0.7804 & 0.7698 & 0.7519 \\
        \hline
    \end{tabular}
    \label{tab:comparative_exp_result_algorithm}
\end{table}

From the experimental results, it can be seen that there are obvious differences in the performance of different machine learning models in the credit default prediction task. Traditional models such as logistic regression and decision trees have relatively weak overall performance in terms of accuracy, precision and recall. Although logistic regression has certain advantages in terms of accuracy, its overall prediction ability is not as good as that of ensemble learning methods. The performance of KNN and naive Bayes is relatively general, especially in terms of precision and recall, which shows that their generalization ability is limited when processing complex high-dimensional data. In contrast, MLP (multi-layer perceptron)improves the model effect through nonlinear fitting, but it is still slightly inferior to the optimized ensemble algorithm.

The experimental results show that XGBoost, LightGBM, and CatBoost achieve the best overall performance across evaluation metrics, including accuracy, precision, and recall. This finding shows that gradient boosting-based ensemble methods can capture complex, non-linear relationships between features, while maintaining robustness against unbalanced data. Notably, XGBoost outperforms the others in recall, demonstrating its sensitivity in detecting defaulting customers. This advantage makes it well-suited for the credit default predicting task, where minimizing false negatives - that is, minimizing missed defaults - is crucial. Furthermore, XGBoost is recognized as a highly efficient and scalable gradient boosting decision tree, and has shown promising performance in prediction accuracy across various fields\cite{zhong2025enhancing, wang2025systematic}. These findings proved the effectiveness of gradient boosting-based ensemble methods in credit default prediction, highlighting XGBoost as a reliable and scalable choice for credit default prediction models.

Furthermore, this paper also explores the impact of the maximum depth on the model. The experimental results are shown in Table 2.

\begin{table}[h!]
    \centering
    \caption{Model Performance Across Maximum Depths.}
    \begin{tabular}{|c|c|c|c|}
        \hline
        \textbf{Max\_depth} & \textbf{ACC} & \textbf{Precision} &
        \textbf{Recall} \\
        \hline
        3 & 0.7687 & 0.7523 & 0.7368 \\
        4 & 0.7804 & 0.7698 & 0.7519 \\
        5 & 0.7772 & 0.7651 & 0.7484 \\
        6 & 0.7745 & 0.7613 & 0.7447 \\
        7 & 0.7718 & 0.7580 & 0.7402 \\
        \hline
    \end{tabular}
    \label{tab:comparative_exp_result_max_depth}
\end{table}

The experimental results show that the performance of the model under different maximum depth (max\_depth) settings is different. The best result occurs when max\_depth is 4, at which time all indicators reach the highest, among which the F1 value is 0.7605, indicating that this parameter setting performs best in balancing model complexity and generalization ability. In contrast, when the depth is too shallow (such as max\_depth=3), the model accuracy and recall rate both decrease, indicating that the model may be underfitting under this setting.

With the increase of depth (5, 6, 7), although the model improves the fitting ability to a certain extent, the performance indicators show a gradual downward trend, especially in the recall rate and F1 value, indicating that the model begins to overfit. This shows that in the current task, the overly deep tree structure has not brought about a continuous improvement in performance, but may increase the dependence on the training set and affect the generalization ability.

In summary, max\_depth has a significant impact on model performance. The appropriate depth (such as 4) can not only effectively capture the relationship between features and target variables, but also avoid the overfitting problem caused by excessive model complexity. Therefore, max\_depth=4 can be used as a more robust parameter choice in this task.
Finally, this paper also uses Shap for analysis. First, the feature importance analysis is given, as shown in Figure 2.

\begin{figure}[H]
    \centering
    \includegraphics[width=0.5\textwidth]{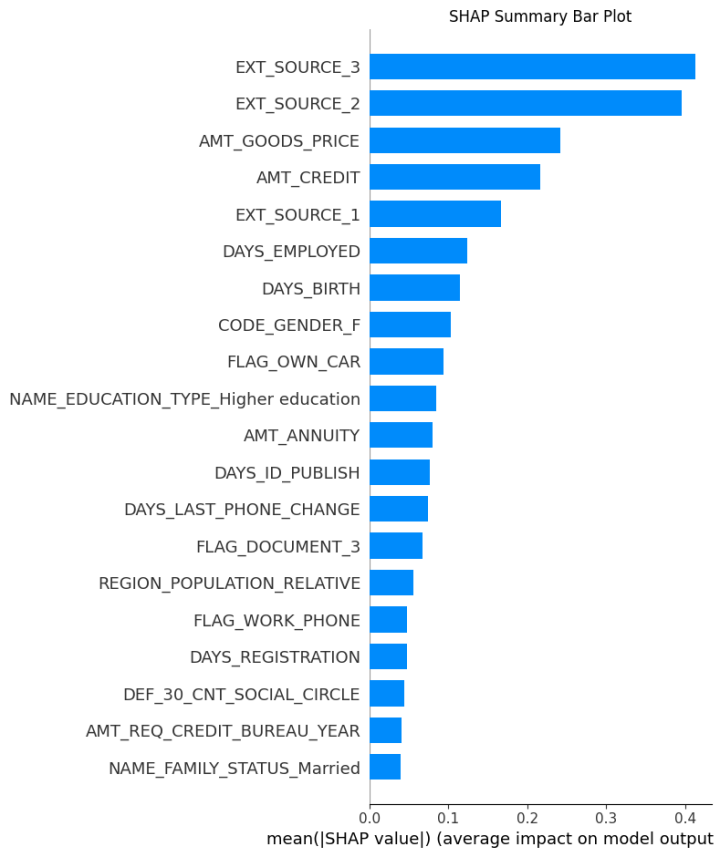}
    \caption{Feature Importance Analysis.}
    \label{fig:featureimportance}
\end{figure}

From the distribution of SHAP values in the figure, it can be seen that the features that the model relies on most are EXT\_SOURCE\_3 and EXT\_SOURCE\_2. Their average impact on the prediction results is much higher than other variables, indicating that these two features play a key role in determining the customer's default risk. These two variables are usually external scoring sources, which can better reflect the customer's credit performance and historical behavior, and have a positive contribution to the stability and generalization ability of the model. In addition, AMT\_GOODS\_PRICE and AMT\_CREDIT are also of high importance, showing that the loan amount and the corresponding commodity price are significantly associated with the probability of default.

Among the remaining features, variables related to the basic information of customers such as DAYS\_EMPLOYED, DAYS\_BIRTH, and CODE\_GENDER\_F still have a certain impact on the model, but it is relatively weak. The influence of factors such as education type, whether there is a car, annuity, and telephone replacement time is moderate, which may be related to the borrower's socioeconomic status and stability, while the influence of features such as telephone, document mark, and regional population density tends to be marginal. This shows that the model mainly relies on external scores and credit amount variables for risk judgment, while its reliance on some traditional static variables is relatively low, further highlighting the value of refined credit scoring models in real-world applications.
Furthermore, this paper also gives the relevant dependency graph, as shown in Figure 3.

\begin{figure}[H]
    \centering
    \includegraphics[width=0.5\textwidth]{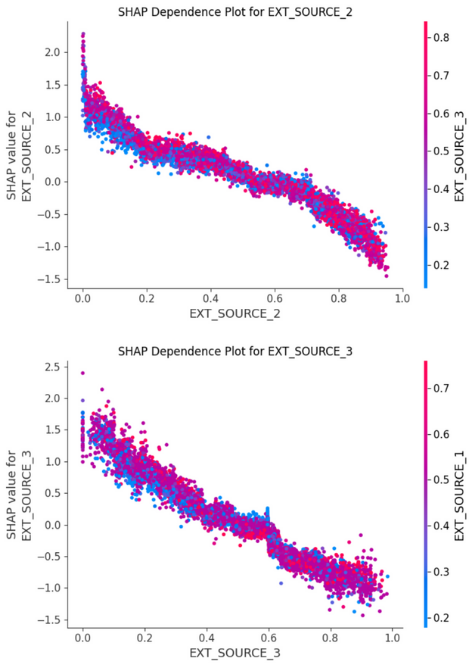}
    \caption{SHAP Dependency Graph.}
    \label{fig:shapdep}
\end{figure}

From the SHAP dependency graph, we can see that the two features EXT\_SOURCE\_3 and EXT\_SOURCE\_2 have a significant negative correlation with the model prediction results. As the values of these two variables increase, their corresponding SHAP values gradually decrease, indicating that the higher the customer's external score, the more the model tends to judge that their default risk is lower. This downward trend shows a relatively stable nonlinear characteristic, especially in the medium and high score range, the change of SHAP values tends to be moderate, reflecting that the model's risk assessment of customers with good credit is more conservative and stable.

In addition, the two dependency graphs also reflect the interactive relationship with their covariates through color: in the EXT\_SOURCE\_3 graph, the color represents EXT\_SOURCE\_1, and in the EXT\_SOURCE\_2 graph, the color represents EXT\_SOURCE\_3. It can be observed that under the same main variable value conditions, the high and low covariates still have a certain regulatory effect on the SHAP value, especially in the low score segment. The impact is more significant, indicating that the model not only focuses on a single score variable when making predictions, but also comprehensively considers the coupling relationship between multiple score features. This also further explains why these three EXT\_SOURCE scores are always at the core of the feature importance ranking. 

\section{Future Development}
While this study addresses the class imbalance problem in the Home Credit dataset using SMOTE, another problem - cross-domain variability - remains challenging. Cross-domain variability is introduced because sub-tables represent features differently. These discrepancies make it difficult for models to learn consistent patterns across domains, affecting their performance and robustness. In future work, adopting unsupervised domain adaptation techniques, such as the relative entropy regularization and measure propagation method proposed by Tan et al. \cite{tan2025unsupervised}, could potentially bridge the gap in feature distribution, enhancing models' ability to generalize across multi-source data in credit default prediction. Additionally, data pruning techniques could be explored as a complement strategy to SMOTE for handling large and imbalanced dataset like Home Credit. Li et al. demonstrated that selectively removing less informative samples can improve training efficiency while preserving model robustness \cite{li2023less,li2024nas,li2024pruning}.

\section{Conclusion}
Through the comparative analysis of various machine learning models in the credit default prediction task in this study, it can be found that ensemble learning methods (such as XGBoost, LightGBM and CatBoost) perform well in core indicators such as accuracy, precision and recall, and are significantly better than traditional models, reflecting their modeling advantages in high-dimensional and nonlinear data environments. At the same time, SHAP feature importance and dependency analysis further reveals the key variables behind the model decision, especially multiple external credit score features, which have a stable and strong impact on the model output, verifying the effectiveness of external data fusion in credit risk control scenarios.

Overall, this study not only constructs a credit default prediction framework based on machine learning, but also systematically verifies the performance differences of different algorithms and the risk indication capabilities of key features. The research results have practical application value for the automation and refinement of the credit approval process, and also provide theoretical support and practical basis for subsequent optimization in the directions of model interpretability, data enhancement and imbalance processing.

\bibliography{references}
\bibliographystyle{ieeetr}
\end{document}